\documentclass[twocolumn]{article}
\usepackage{arxiv}

\usepackage[utf8]{inputenc}
\usepackage[T1]{fontenc}
\usepackage{microtype}

\usepackage{amsmath,amssymb,bm}
\usepackage{amsfonts}
\usepackage{nicefrac}
\usepackage{enumitem}
\usepackage[table]{xcolor}
\usepackage{graphicx}
\graphicspath{{./}{Figures/}{figures/}}
\usepackage{tikz}
\usepackage{pifont}

\usepackage{booktabs}
\usepackage{array}
\usepackage{makecell}
\usepackage{multirow}
\usepackage{multicol}
\usepackage{longtable}
\usepackage{caption}
\usepackage{threeparttable}

\usepackage{natbib}
\usepackage{authblk}
\usepackage{url}
\usepackage{hyperref}
\usepackage{doi}

\makeatletter
\renewcommand{\maketitle}{%
  \par
  \begingroup
    \renewcommand{\thefootnote}{\fnsymbol{footnote}}
    \renewcommand{\@makefnmark}{\hbox to \z@{$^{\@thefnmark}$\hss}}
    \long\def\@makefntext##1{%
      \parindent 1em\noindent
      \hbox to 1.8em{\hss $\m@th ^{\@thefnmark}$}##1
    }
    \twocolumn[\@maketitle]%
    \thispagestyle{empty}
    \@thanks
  \endgroup
  \let\maketitle\relax
  \let\thanks\relax
}
\makeatother
\AtBeginDocument{\setlength{\columnsep}{0.25in}}

\renewenvironment{abstract}{%
  \centerline{\large\bfseries\scshape Abstract}%
  \vspace{0.4em}\par\noindent\ignorespaces
}{%
  \par\vspace{0.8em}%
}

\setcitestyle{numbers,square,comma} 
\setcellgapes{5pt}

\newcommand{\cmark}{\textcolor{teal}{\ding{51}}}
\newcommand{\xmark}{\textcolor{red}{\ding{55}}}

\title{VOLA: Improving \textcolor{blue}{O}pen-World Driving by \textcolor{blue}{VL}M-Based \\Semantic \textcolor{blue}{A}ttribute Prediction}

\renewcommand{\shorttitle}{VOLA: Improving Open-World Driving by VLM-Based Semantic Attribute Prediction}

\author[1]{Yuchen Zhang}
\author[1]{Yuan Gao}
\author[2]{Sebastian Schmidt}
\author[1]{Johannes Betz}

\affil[1]{Professorship of Autonomous Vehicle Systems, Technical University of Munich, Munich, Germany}
\affil[2]{Data Analytics and Machine Learning Group, Technical University of Munich, Munich, Germany \protect\\ Contact: \{yuchen2.zhang, yuan\_avs.gao, sebastian95.schmidt, johannes.betz\}@tum.de}

\date{}

\begin{document}
\maketitle
\begin{abstract}
Driving in the real world is open-world: a car may encounter a fallen mattress, a deer, or other objects outside its training data.
Naming them is not enough.
The system must know how to treat each region: can it drive over it, how severe would a collision be?
We therefore shift scene perception from category labels to dense action-relevant attributes, where each pixel is labeled by how it should affect motion rather than by object name.
We instantiate this general formulation with two ordered attributes: 7-rank drivability and 5-rank vulnerability.
We read Qwen3.5 image-token hidden states directly as a spatial semantic representation.
A lightweight boundary-aware decoder then turns this coarse token grid into sharp full-resolution attribute maps.
The whole process requires \textit{neither} autoregressive text generation \textit{nor} an external mask model such as SAM.
We train on dense attribute labels built in CARLA and test transfer to real scenes and to novel obstacles never seen in training. 
We compare with vision-only segmenters trained on the same attributes and prompted VLM segmenters.
Our model matches strong vision-only segmenters on familiar categories and improves transfer to real open-world anomalies, reaching 69.4\% mean vulnerability-rank recall versus 57.1\% for the best vision-only baseline and 53.9\% for the best prompted VLM
baseline.
These results show that VLM image tokens provide useful semantic cues for transferring driving attributes to objects outside the training vocabulary. 
Code is available \href{https://anonymous.4open.science/r/VOLA}{here}.
\end{abstract}
    
\section{Introduction}
\label{sec:intro}

Autonomous driving in the real world is fundamentally an open-world problem, where rare objects and unknown hazards may appear during deployment~\cite{pinggera2016lostandfound,chan2021segmentmeifyoucan}.
A mattress that fell from a truck, a loose tire on the highway, or a deer crossing a rural road may each be rare, but such long tail cases are collectively unavoidable.
Handling them safely is one of the main remaining obstacles to reliable autonomy~\cite{koopman2016challenges, makansi2021exposing}.

The first bottleneck is the closed-set nature of standard semantic segmentation: every pixel must be assigned to one class from a label set fixed before training~\cite{xiao2018upernet,xie2021segformer,cheng2022mask2former}.
This interface works when the scene fits the annotation taxonomy, but it has no reliable output for an object outside the label set. The model must either ignore the region, absorb it into background, or force it into the nearest known class.

Open-set and open-world perception address this failure by adding a discovery mechanism.
Instead of forcing every pixel into a known class, these methods use signals such as energy scores~\cite{owod,tian2022pixel}, objectness priors~\cite{kim2022learning,zohar2023prob}, or evidential
uncertainty~\cite{ancha2024deep,schmidt2025prior2former} to flag regions outside the known label set.
They separate regions into known classes and a generic "unknown" bucket.
This reduces silent failure, but the system still does not know what to do with that region.
An ``unknown'' label does not say whether the region is safe to enter, costly to hit, or occupied by a vulnerable agent.

A different route is to draw on vision-language models (VLMs), whose large-scale image-text training provides broader visual knowledge~\cite{bai2025qwen3,chen2024internvl}.
Recent work \cite{jin2024tod3cap,inoue2024nuscenes} has applied VLMs to driving scenes for captioning, where they can produce rich descriptions of what they observe.
However, simply reporting what the model sees, calling a tire ``tire'' and a mattress ``mattress'', is not enough.
Planning operates on costs and constraints, such as whether a region can be entered and whether a collision would be severe.
For common categories such as cars, pedestrians, cyclists, and traffic signs, these meanings are usually already encoded in the driving stack.
For rare or unseen categories, they are not.
If perception reports a ``tire''or a ``mattress'', the planner still needs to know how that region should affect motion.
The system must either maintain a class-to-cost mapping for the long tail of objects that may appear during deployment, which is unrealistic, or learn this mapping from sparse long-tail data.
\textit{Shifting from category-centered perception to attribute-centered perception could remove this extra step by describing how each region matters for driving rather than only naming what is there.}

We therefore recast driving-scene perception as dense ordinal attribute prediction.
\textit{Instead of assigning each pixel a visual category, we predict per-pixel ranks for driving-relevant properties such as drivability and vulnerability.}
To predict these maps, we use Qwen3.5~\cite{qwen35blog} as a dense semantic source rather than as a text generator.
Its image token hidden states form a coarse spatial grid, preserving broad image-text knowledge in a localized representation.
A lightweight boundary-aware decoder turns this grid into full resolution attribute rank maps, without relying on an external promptable segmentation model such as SAM~\cite{kirillov2023sam}.

In summary, our contributions are:
\begin{itemize}[noitemsep, topsep=0pt, leftmargin=*]

  \item We formulate open-world driving perception as dense attribute prediction, shifting the output space from fixed object classes to driving-relevant properties.

  \item We show with VOLA that image tokens can serve as a dense semantic source, allowing attribute prediction without text generation or special segmentation tokens.

  \item We build dense attribute supervision in CARLA simulator, and show that the learned attributes generalize beyond the simulator to real scenes and beyond the training taxonomy to unseen open-world obstacles.
\end{itemize}

\section{Related Work}
\label{sec:related}

\paragraph{Closed-set segmentation.}
Semantic segmentation is a fixed-label dense prediction task: given an image and a label set, predict a class for every pixel.
Fully convolutional networks made the task end-to-end by turning image-level classifiers into dense predictors~\cite{long2015fcn}.
Later work strengthened the dense representation with context and multi-scale reasoning: pyramid pooling in PSPNet~\cite{zhao2017pspnet}, atrous encoder-decoder features in DeepLabv3+~\cite{chen2018deeplabv3plus}, and multi-level parsing in UPerNet~\cite{xiao2018upernet}.
Transformer and mask-classification methods improved quality further: SegFormer~\cite{xie2021segformer} pairs a hierarchical transformer encoder with a light decoder, MaskFormer~\cite{cheng2021maskformer} recasts segmentation as mask classification, and Mask2Former~\cite{cheng2022mask2former} extends that interface across semantic, instance, and panoptic segmentation.

All of these approaches can only predict classes induced in their fixed training vocabulary.
Content outside the vocabulary is absorbed into the background, ignored, or forced into the nearest known class.

\vspace{-5pt}
\paragraph{Open-world detection and segmentation.}

Open-world methods relax the fixed vocabulary by flagging pixels or regions that do not fit any known class.
They differ mainly in where the unknown signal comes from.
Some methods use model-internal confidence~\cite{jung2021standardized}, energy~\cite{owod}, or evidential uncertainty~\cite{schmidt2025prior2former} scores to reject regions outside the known classes.
Others derive unknown supervision from the training data itself, for example, by mining object-like regions that do not match known annotations~\cite{gupta2022ow}. Other methods learn category-agnostic objectness from known instances~\cite{kim2022learning,zohar2023prob} or use external negative data to strengthen unknown detection~\cite{chan2021entropy,tian2022pixel,grcic2022densehybrid}.

While those methods are able to reduce silent failure on out-of-taxonomy regions, the discovery stops at an ``unknown'' flag.
Knowing that a region is ``unknown'' says nothing about how to treat it: a deer and a fallen mattress may both be unknown, yet one is vulnerable and the other is an obstacle.

\vspace{-5pt}
\paragraph{Open-vocabulary and VLM-based segmentation.}
Open-vocabulary and VLM-based segmentation offer a different route to the open world. The target there is specified in language rather than fixed by a label index.
Existing methods differ in which representation conditions the mask.
CLIP-style open-vocabulary methods condition on class-name embeddings: OVSeg~\cite{ovseg} scores class-agnostic mask proposals against CLIP \cite{clip} text embeddings, and CAT-Seg~\cite{cho2024catseg} builds dense CLIP features and matches them to class names per pixel.
Instruction-tuned segmenters condition on language-side tokens instead.
LISA~\cite{lai2024lisa} prompts SAM \cite{kirillov2023sam} with the hidden state of one generated segmentation token, PixelLM~\cite{ren2024pixellm} feeds learned segmentation tokens to a pixel decoder, and GSVA~\cite{xia2024gsva} adds empty-target rejection.
F-LMM~\cite{fllm} grounds words of an ordinary assistant response through attention maps before mask decoding and SAM refinement, while PSALM~\cite{zhang2024psalm} appends learnable mask tokens to the input and decodes masks from their output embeddings.
Table~\ref{tab:vlm-seg-compare} summarizes the main design choices along which these methods differ.

While these methods make segmentation more flexible by conditioning masks on language, the output remains tied to a queried concept or generated segmentation token.
A prompt such as ``mattress'' or ``deer'' can localize the object, yet the resulting mask still does not specify how the region should matter for driving, which we address in our work.

  \begin{table}[htbp]
    \centering
    \footnotesize
    \setlength{\tabcolsep}{2pt}
    \renewcommand{\arraystretch}{1}
    \begin{tabular}{@{}lcccc@{}}
    \toprule

     & \shortstack{No\\special tok.} & \shortstack{No autoreg.\\gen.}
     & \shortstack{Without\\SAM 
     \cite{kirillov2023sam}} & \shortstack{Reads\\img. tok.} \\
    \midrule
    LISA~\cite{lai2024lisa}       & \xmark & \xmark & \xmark & \xmark \\
    PixelLM~\cite{ren2024pixellm} & \xmark & \xmark  & \cmark & \xmark\\
    GSVA~\cite{xia2024gsva}       & \xmark & \xmark & \xmark & \xmark \\
    PSALM~\cite{zhang2024psalm}   & \xmark & \cmark & \cmark  &\xmark \\
    F-LMM~\cite{fllm}             & \cmark & \xmark & \xmark & \xmark \\
    \textbf{VOLA [ours]}          & \cmark & \cmark & \cmark & \cmark \\
    \bottomrule
    \end{tabular}
    \caption{\textbf{VLM-based segmenters.}
  Unlike prior methods, VOLA reads image tokens directly, without special tokens, text generation, or SAM.}
    \label{tab:vlm-seg-compare}
  \end{table}

\begin{figure*}[htbp]
\centering
\includegraphics[width=\textwidth]{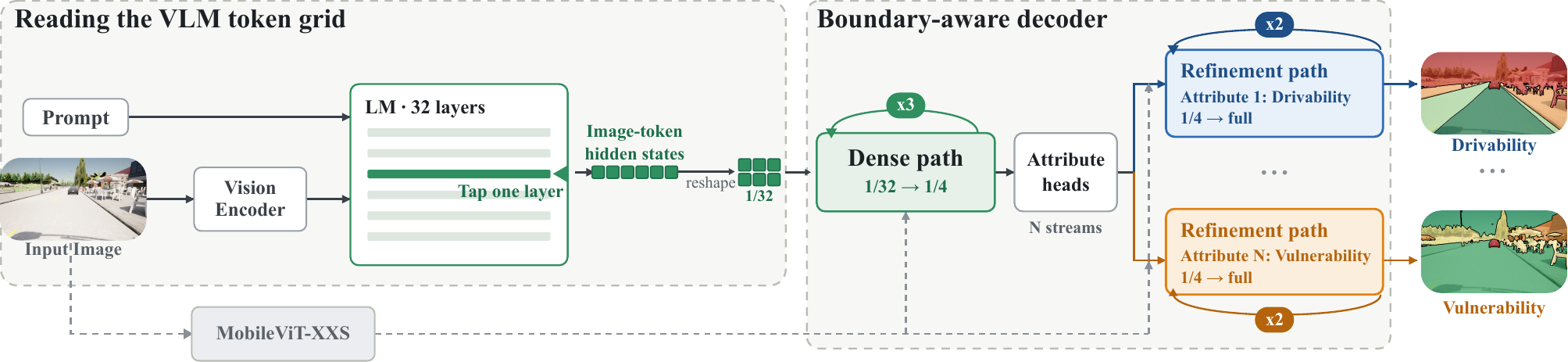}
\caption{\textbf{Method overview.}
Given an image and a short prompt, we run the VLM once and \textit{read image-token hidden states} from an intermediate layer, without text generation or added special tokens. The tokens are reshaped into a $1/32$ spatial
grid, which provides semantic scene features but is too coarse for accurate boundaries. A \textit{boundary-aware decoder} upsamples this grid to $1/4$ resolution while fusing RGB features from a lightweight MobileViT-XXS \cite{mehta2022mobilevit} branch. Attribute
heads then split the shared feature map into per-attribute streams. Each stream is refined from $1/4$ to full resolution at uncertain points, producing one dense rank map per driving attribute.}
\label{fig:pipeline}
\end{figure*}

\section{Method}
\label{sec:method}

VOLA addresses two gaps in dense perception for open-world driving.
First, closed-set segmenters and open-world methods still describe regions through object classes or unknown flags.
Neither output says how the region should affect driving.
We therefore predict ordered driving attributes, such as drivability and vulnerability.
Second, VLM-based segmenters usually localize queried concepts through generated text, special tokens, or external mask decoders.
VOLA instead reads image-token hidden states directly and treats them as a spatial semantic field.
Figure~\ref{fig:pipeline} shows how this field is decoded: the image-token grid is reshaped, upsampled by a boundary-aware decoder, split into attribute streams, and refined to full-resolution rank maps.

\subsection{Problem formulation}
\label{subsec:formulation}

Let $I \in \mathbb{R}^{3 \times H \times W}$ be an input image.
Given a set of ordered attributes $\mathcal{A}$, VOLA predicts one dense rank map for each attribute.
For attribute $a$, ranks take values in $\mathcal{Y}_a=\{0,\ldots,K_a-1\}$.
The final output has size $|\mathcal{A}| \times H \times W$.

In this work, as a concrete instantiation of the general framework, we use two attributes, drivability and vulnerability.
The framework does not depend on these two attributes or on these exact numbers of ranks.
Other tasks can define other ordered attributes.

We define the ranks according to two principles. First, each rank should correspond to a different motion or contact consequence.
Second, the order should be meaningful, so distant mistakes are more severe than nearby mistakes.

Following these principles, drivability measures how suitable a region is as a motion target for the ego vehicle. 
Higher ranks mean better motion targets.
Rank 6 denotes normal motion in the current lane and its forward continuation.
Rank 5 denotes a legal same-direction lane change target.
Rank 4 denotes same direction road that is not legally reachable from the current lane.
Rank 3 denotes road that would otherwise be usable but is currently blocked by a red light.
Rank 2 denotes an emergency off road fallback.
Rank 1 denotes an opposite direction lane.
Rank 0 denotes no valid motion target, including occupied regions and non-driving regions such as sky.

Vulnerability measures how severe the consequences of contact with a region would be.
Higher ranks mean higher contact cost.
Rank 4 denotes unprotected biological agents, such as pedestrians, cyclists, and animals.
Rank 3 denotes high-cost contact with protected agents, such as cars, trucks, and buses.
Rank 2 denotes heavy rigid structures such as wall and buildings.
Rank 1 denotes light obstacles and roadside structures.
Rank 0 denotes regions where there is no object to collide with.
  
\subsection{Reading the VLM token grid}
\label{subsec:representation}

Predicting drivability and vulnerability needs semantic knowledge about objects, scenes, and what they mean for driving.
VLMs hold this knowledge from large-scale image-text pretraining, but they usually expose it as generated text, whereas dense attributes need a spatial form.
This form actually already exists inside the model before text generation, since the VLM encodes the image as a grid of token features with one vector per region.
We read this grid directly and use it as our spatial semantic field.

This keeps the method simple and cheap.
It needs only a single forward pass, with no autoregressive text generation.
It also adds no special tokens, since the image tokens the model already produces are exactly what we read.

We give the model a short text prompt of $m$ tokens followed by the image.
The visual encoder splits the image into a $g_h \times g_w$ patch grid.
It then merges every $s \times s$ block into one image token, giving $n = \tfrac{g_h g_w}{s^2}$ image tokens.
During visual encoding, image tokens exchange information with other image tokens.
Thus, each token carries local appearance and image context before it enters the language model.

The VLM then processes the prompt tokens and image tokens as one sequence,

\begin{equation}
X = (\,\underbrace{p_1,\dots,p_m}_{\text{prompt}},\ \underbrace{v_1,\dots,v_n}_{\text{image}}\,).
\end{equation}
Each transformer layer can enrich the content of each token, but it does not change the number or order of image token positions.
The hidden state at an image token position can therefore carry context from the image and the prompt, while still corresponding to a fixed region of the input image.
We then select the $n$ image token hidden states and reshape them into an image-aligned grid of $\tfrac{g_h}{s} \times \tfrac{g_w}{s}$ cells.

More concretely, we read image token hidden states from layer~19 of~32, selected by the ablation in Sec.~\ref{ablation:tapped layer}.
Qwen3.5 uses $16\times16$ patches with a $2\times2$ merge.
Each cell therefore spans a $32\times32$ image region, giving about $\tfrac{H}{32}\times\tfrac{W}{32}$ cells for an $H\times W$ image.

\subsection{Boundary-aware decoder}
\label{subsec:decoder}

The VLM grid is semantically rich but spatially coarse, with each token summarizing a $32\times32$ image region.
Directly projecting this grid to full resolution would blur thin structures and object boundaries.
We therefore use a boundary-aware decoder that preserves the VLM semantic signal while recovering missing spatial detail from the image.
It upsamples the grid through two successive paths: a dense path that progressively fuses RGB appearance features, and a refinement path that sharpens predictions at still-uncertain pixels, which typically lie near object boundaries.

The dense path upsamples the VLM token grid in three stages, from $1/32$ to $1/4$ resolution, while fusing RGB features from a lightweight MobileViT-XXS encoder~\cite{mehta2022mobilevit}.
At each scale, the RGB fusion branch is zero-initialized, so the decoder starts from the VLM semantic signal and learns to add local appearance cues only when useful.
RGB features, therefore, restore detail for thin structures and object boundaries without taking over the semantic prediction.
Finally, at $1/4$ resolution, each attribute has its own head: a $1\times1$ convolution that maps the decoder feature at every pixel $p$ to coarse logits $z_a(p) \in \mathbb{R}^{K_a}$, one entry per rank.

The refinement path follows PointRend~\cite{kirillov2020pointrend} and refines uncertain points, which are often concentrated near thin structures and object boundaries.
For each selected point $p$, a small MLP predicts refined logits $\hat{z}_a(p)$ from the coarse logits $z_a(p)$ and a fine image feature sampled at the same location.
We apply this refinement in two $\times2$ stages, lifting the prediction from $1/4$ resolution to full resolution.
During training, we sample a fixed point budget biased toward uncertain locations, while at inference, we refine the most uncertain points at each upsampling stage.
We ablate this refinement against bilinear upsampling in Sec.~\ref{ablation:pointrend}.

Since each attribute label is ordered, the loss should reflect rank distance.
For example, predicting drivability rank 5 instead of rank 6 is a smaller error than predicting rank 0.
Ordinal losses such as CORAL~\cite{cao2020coral} and CORN~\cite{shi2023corn} encode this ordering by decomposing rank prediction into ordered binary decisions.
They are, therefore, natural choices for our attributes.
However, we find that a simple sigmoid focal loss performs better in practice (Sec.~\ref{ablation:loss}).
The final training contains a dense term on the coarse logits and a sparse term on the refined points:

\begin{equation}
\begin{split}
\mathcal{L} = \sum_{a \in \mathcal{A}} \bigg[
  &\sum_{p \in \Omega_c} \ell\big(z_a(p), y_a(p)\big) \\
  &+ \sum_{p \in P} \ell\big(\hat{z}_a(p), y_a(p)\big) \bigg],
\end{split}
\end{equation}
where $\ell$ is the focal loss, $y_a(p)$ is the ground-truth rank, $\Omega_c$ is the pixel grid of the coarse prediction, and $P$ is the set of points selected for refinement.
At inference, each pixel takes the highest-scoring rank on each attribute.
\section{Dataset Construction}
\label{sec:dataset}

Our model needs dense per-pixel attribute supervision.
We considered building such supervision from existing real-world segmentation datasets such as Cityscapes~\cite{cordts2016cityscapes} and nuImages~\cite{caesar2020nuscenes} by mapping each annotated class to a fixed attribute combination, but this approach has two limitations.

First, their object taxonomies are fixed and incomplete, so out-of-distribution objects are often unlabeled or absorbed into broad fallback classes such as ``static'' and ``dynamic''.
Second, the \texttt{road} class is itself too coarse.
It alone covers $32.6\%$ of all pixels in Cityscapes and $21.2\%$ in nuImages, but it merges regions that should receive different attribute labels.
For example, the ego lane is currently drivable by the ego vehicle, a bicycle lane is reserved for cyclists, and an oncoming lane belongs to traffic moving in the opposite direction.
A single \texttt{road} label cannot distinguish these cases.
As a result, class-to-attribute conversion would be incomplete for some objects and ambiguous for road structure.

\paragraph{Collection in CARLA.}
We instead collect data in the CARLA simulator~\cite{dosovitskiy2017carla}, where the class of every object and lane connectivity are directly available.
The ego vehicle drives under autopilot through CARLA's Traffic Manager, with traffic and pedestrians populating the scene under continuously varying weather.
Frames are sampled every two seconds of simulated time, and any frame whose ego pose has moved less than $0.1$ m from the previous saved frame is discarded to remove near-duplicates from idling at signals or in dense traffic.

\paragraph{Label construction.}
Labels are built from three sources.
\emph{(i) Lane structure} comes from the CARLA map.
Starting at the ego's waypoint, we trace the lane network forward through successor waypoints, branching at intersections, and label each visible road pixel by its connectivity to the ego: the current lane and its forward continuation, lanes reachable through legal lane changes and their forward continuations, same-direction lanes that are not reachable, or oncoming lanes.
\emph{(ii) Temporary lane availability} is added on top of the lane structure map.
This context marks lanes in the ego's travel direction that are temporarily unavailable because of a red light.
\emph{(iii) Object semantics} come from CARLA's segmentation camera.

\paragraph{Splits.}
We split the CARLA data by town instead of randomly splitting frames.
A random frame split would create leakage between the training and test sets because each CARLA town covers a limited road network.
Since the autopilot naturally revisits the same streets during data collection, repeated views of the same road segments could appear in both splits.
A town-level split gives a stricter evaluation, since each held-out town contains road layouts that the model has not seen during training.
The training set contains Town02, Town03, Town04, and Town05, for a total of $3{,}752$ frames.
The validation set contains Town01, with $736$ frames, and the test set contains Town10HD, with $200$ frames.

\section{Experiments}
\label{sec:experiments}

Our experiments test the central claim that VLM image tokens can support dense driving-attribute prediction beyond the training vocabulary.
We ask two questions.
\textit{First, does a VLM backbone improve attribute prediction compared with strong vision-only segmenters trained on the same labels?}
\textit{Second, is attribute supervision necessary, or can existing VLM segmenters produce these maps by prompting alone?}
We answer these questions under progressively stronger distribution shifts: from a style-shifted CARLA town, to real-world Cityscapes \cite{cordts2016cityscapes} images, to synthetic novel-object anomalies, and finally to real novel-object anomalies.

\subsection{Datasets and Metrics}
\label{sec:eval_datasets}

We train on the CARLA training split then evaluate under two kinds of shift: visual shift and semantic novelty.

For visual shift, we evaluate on the CARLA validation split, the CARLA test split, and Cityscapes.
The CARLA test split is collected in Town10HD, which is a special CARLA town with a modern downtown layout and a visual style that is clearly different from the training towns.
Cityscapes moves from simulation to real driving images.
All datasets provide dense labels, so we report per-axis mIoU.
Since Cityscapes has no lane connectivity or traffic-rule labels, we collapse drivability into three groups: non-drivable, off-road, and on-road, while keeping vulnerability at five ranks.

For semantic novelty, we evaluate on StreetHazards \cite{hendrycks2019anomalyseg} and SegmentMeIfYouCan (SMIYC) \cite{chan2021segmentmeifyoucan} AnomalyTrack.
StreetHazards places unseen objects into the CARLA simulator.
SMIYC contains real-world anomalous objects such as animals and lost cargo.
StreetHazards and SMIYC label only anomaly pixels, not the full scene, so we report anomaly-pixel recall.
For drivability, recall means predicting the anomaly as non-drivable.
For vulnerability, recall means predicting the correct manually annotated rank.

\subsection{Experiment 1: Is a VLM backbone necessary?}
\label{sec:exp1}

Modern vision-only segmenters are strong dense predictors, and our attribute maps are dense prediction targets. It is therefore possible that, when trained with the same attribute supervision, a conventional segmenter can already
learn the appearance, geometry, and scene-context cues needed to infer how each region should be treated. Experiment 1 tests this alternative by comparing VOLA with standard dense segmenters under identical supervision,
asking whether intermediate VLM image-token features offer stronger transfer than vision-only features as evaluation shifts from familiar scenes to novel objects.

\vspace{-5pt}
\paragraph{Baselines.}
We use four standard segmenters that span convolutional and transformer designs:
DeepLabV3+~\cite{chen2018deeplabv3plus}, UperNet~\cite{xiao2018upernet},
SegFormer~\cite{xie2021segformer}, and Mask2Former~\cite{cheng2022mask2former}.
These models are designed to produce a single label map. To produce both
attributes while leaving their architectures untouched, we train a separate
model for each, one for drivability and one for vulnerability. All baselines use
their standard ImageNet-pretrained backbones and default training recipe, and
are trained on the same CARLA images, attribute labels, and splits as our model.

  \begin{table*}[t]
  \centering
  \renewcommand{\arraystretch}{1.3}
  \setlength{\tabcolsep}{4.5pt}
  \scriptsize
  \begin{threeparttable}
  \caption{\textbf{Comparison with vision-only segmenters.}
  CARLA and Cityscapes report per-axis mIoU.
  StreetHazards and SMIYC report vulnerability-rank recall on anomaly pixels.
  \textbf{Bold}: best, \underline{underline}: second best.}
  \label{tab:exp1}
  \begin{tabular}{@{}l rrrrrr @{\hspace{1.5em}} rrrrr rrrr@{}}
  \toprule
  & \multicolumn{6}{c}{mIoU (\%) $\uparrow$} & \multicolumn{9}{c}{Novel-object vulnerability recall (\%) $\uparrow$} \\
  \cmidrule(lr){2-7} \cmidrule(lr){8-16}
  & \multicolumn{2}{c}{CARLA val} & \multicolumn{2}{c}{CARLA test} & \multicolumn{2}{c}{Cityscapes} & \multicolumn{5}{c}{StreetHazards} & \multicolumn{4}{c}{SMIYC} \\
  \cmidrule(lr){2-3}\cmidrule(lr){4-5}\cmidrule(lr){6-7}\cmidrule(lr){8-12}\cmidrule(lr){13-16}
  Model & driv & vul & driv & vul & driv & vul & rank 1 & rank 2 & rank 3 & rank 4 & mean & rank 1 & rank 3 & rank 4 & mean \\
  \midrule
  SegFormer \cite{xie2021segformer}    & \underline{80.80} & 87.03 & \underline{79.50} & 80.82 & 81.21 & 77.46 & 59.96 & 73.64 & 23.65 & 4.98 & 40.56 & 17.73 & 83.83 & \underline{69.58} & \underline{57.05} \\
  UperNet \cite{xiao2018upernet}     & 75.63 & \textbf{91.15} & 74.85 & \textbf{85.45} & 81.27 & \underline{80.73} & 64.59 & \textbf{86.85} & \underline{44.27} & 39.21 & \underline{58.73} & \underline{25.84} & 83.04 & 44.99 & 51.29 \\
  DeepLabV3+ \cite{chen2018deeplabv3plus}  & 80.41 & 85.69 & 74.02 & 72.84 & 68.19 & 74.23 & 62.04 & 75.14 & 36.21 & \underline{48.10} & 55.37 & 8.09 & 78.32 & 60.00 & 48.80 \\
  Mask2Former \cite{cheng2022mask2former} & 76.51 & \underline{88.79} & 73.50 & \underline{83.97} & \underline{83.16} & \textbf{81.48} & \underline{70.71} & 70.05 & 41.37 & 25.42 & 51.89 & 20.53 & \underline{89.95} & 20.40 & 43.63 \\
  \midrule
  Ours: VOLA         & \textbf{80.87} & 87.15 & \textbf{79.74} & 83.57 & \textbf{84.38} & 80.39 & \textbf{78.94} & \underline{79.22} & \textbf{55.88} & \textbf{54.91} & \textbf{67.24} & \textbf{38.25} & \textbf{92.27} & \textbf{77.52} & \textbf{69.35} \\
  \bottomrule
  \end{tabular}
  \end{threeparttable}
  \end{table*}
  
\vspace{-5pt}
\paragraph{Results.}

Table~\ref{tab:exp1} shows that the benefit of VLM image tokens is small under visual shift but clear under semantic novelty.
The three dense splits (CARLA
val, CARLA test, and Cityscapes) contain familiar object types, although their appearance shifts from the training towns to real
Cityscapes images. They therefore mainly test transfer across visual style. The
vision-only baselines handle this setting well, likely because their
ImageNet-pretrained backbones already provide strong appearance features. Our
model is competitive with them on these splits: it performs best on drivability
and remains close on vulnerability. In other words, when the objects are
familiar, a strong conventional segmenter is often enough.

The two novel-object splits (SMIYC and StreetHazards) are different
because they contain object categories never seen during training, such as horses and excavators. This makes them a test
of semantic generalization, not just visual domain transfer. Here, our model
shows a clear advantage. The largest gains are on the most safety-critical rank 4
(unprotected biological agents such as humans or animals): our model recalls
77.5\% of them on SMIYC and 54.9\% on StreetHazards, compared with at
most 69.6\% and 48.1\% for any baseline. Mask2Former, despite being the strongest
model on familiar scenes, recalls only 20.4\% and 25.4\% in this setting. Thus, strong performance on seen objects does not necessarily translate to
strong performance on novel objects.  In the SMIYC example in Fig.~\ref{fig:qualitative}, UPerNet and Mask2Former assign the correct vulnerability rank to only small parts of the nearby cows and largely ignore the distant cows.
Our model assigns the vulnerable rank more consistently across both nearby and distant objects. For anomaly drivability, the task reduces to detecting that each labeled anomaly is non-drivable.
All methods perform similarly on this binary check, with 97.8--99.4\% recall on SMIYC and 98.7--99.2\% on StreetHazards.

\subsection{Experiment 2: Is training necessary?}
\label{sec:exp2}

Experiment 1 shows that VLM image-token features improve transfer under semantic novelty. This raises a second question: if VLMs already contain broad visual knowledge, can existing VLM-based segmenters recover these maps without
additional training? These methods accept language prompts and produce masks, so one might expect them to obtain our attribute maps by prompting each rank description directly. Experiment 2 tests this zero-shot alternative by
comparing VOLA with prompted VLM segmenters, asking whether prompting alone can yield reliable dense attribute maps without attribute-specific training.

\paragraph{Baselines.}
We compare with VLM-based segmentation methods under a zero-shot protocol, with no attribute-specific fine-tuning. Since these methods take language as input, we prompt each model with the text description of each rank, covering seven drivability
ranks and five vulnerability ranks.

Since existing VLM segmenters expose different interfaces, we adapt their outputs
to a common dense rank-map format. For CLIP-based open-vocabulary methods,
OVSeg~\cite{ovseg} and CorrCLIP~\cite{zhang2025corrclip}, the rank descriptions are used
directly as the class vocabulary, and each pixel is assigned to its best-matching
rank. Generative VLM-based referring methods, LISA~\cite{lai2024lisa},
GSVA~\cite{xia2024gsva}, PSALM~\cite{zhang2024psalm}, and F-LMM~\cite{fllm}, are
designed to answer ``where is X?'', where X is a text description of the target.
We set X to each rank description in turn, stack the returned masks, and select
the highest-scoring rank at each pixel.

\begin{table}[t]
\centering
\setlength{\tabcolsep}{4pt}
\renewcommand{\arraystretch}{1.15}
\scriptsize
\begin{tabular}{@{}lcccccc@{}}
\toprule
 & \multicolumn{2}{c}{mIoU (\%) $\uparrow$} & \multicolumn{4}{c}{Recall (\%) $\uparrow$} \\
\cmidrule(lr){2-3}\cmidrule(lr){4-7}
 & \multicolumn{2}{c}{CARLA test} & \multicolumn{2}{c}{SH} & \multicolumn{2}{c}{SMIYC} \\
\cmidrule(lr){2-3}\cmidrule(lr){4-5}\cmidrule(lr){6-7}
Model & driv & vul & driv & vul & driv & vul \\
\midrule
OVSeg \cite{ovseg}    & 10.41 & 26.75 & 71.83 & 45.97 & 61.75 & 18.91 \\
CorrCLIP \cite{zhang2025corrclip} & 3.77  & 48.23 & 12.92 & 41.95 & 1.35  & 31.42 \\
LISA \cite{lai2024lisa}    & \underline{30.79} & \underline{55.72} & \underline{91.57} & 44.41 & \underline{69.43} & \underline{53.90} \\
GSVA \cite{xia2024gsva}    & 26.46 & 40.14 & 34.83 & 42.70 & 32.18 & 40.27 \\
PSALM \cite{zhang2024psalm}   & 9.95  & 32.32 & 21.80 & 41.13 & 19.92 & 19.69 \\
F-LMM  \cite{fllm}  & 6.58  & 36.33 & 28.23 & \underline{54.60} & 14.34 & 40.56 \\
\midrule
Ours: VOLA     & \textbf{79.74} & \textbf{83.57} & \textbf{99.11} & \textbf{67.24} & \textbf{98.63} & \textbf{69.35} \\
\bottomrule
\end{tabular}
\caption{\textbf{Comparison with prompted VLM segmenters.}  CARLA test reports per-axis mIoU.
  StreetHazards and SMIYC report anomaly-pixel recall.
  \textbf{Bold}: best, \underline{underline}: second.}
\label{tab:exp2}
\end{table}

\begin{figure*}[htbp]
\centering
\includegraphics[width=0.94\textwidth]{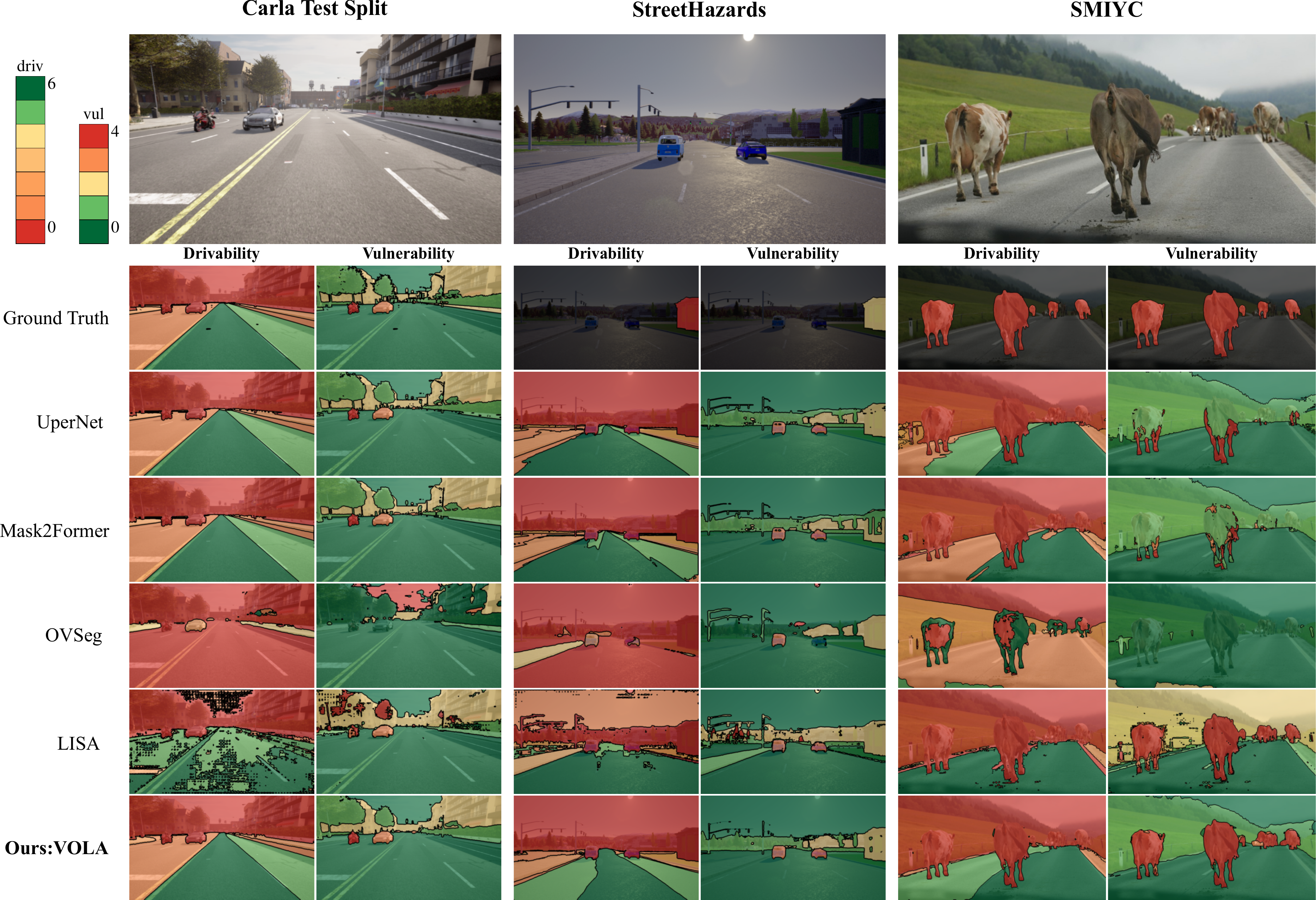}
\caption{\textbf{Qualitative comparison across datasets.}
  Columns follow an increasing distribution shift: CARLA Test changes visual style, StreetHazards introduces synthetic novel objects, and SMIYC introduces real novel objects. Rows compare ground truth, two vision-only segmenters trained on our attribute labels (UPerNet and Mask2Former), two prompted VLM segmenters used without attribute training (OVSeg and LISA), and our model VOLA. CARLA provides dense scene labels, while StreetHazards and SMIYC provide masks only for anomalous objects. Green denotes high drivability or low vulnerability, and red denotes non-drivable regions or highly vulnerable agents. Vision-only models handle CARLA test well but generalize poorly to novel objects, such as the distant cows in SMIYC. Prompted VLM segmenters show clear limitations. OVSeg often assigns one rank to most of the scene. LISA usually captures the coarse vulnerability level, but its maps remain noisy. For drivability, it often merges different road areas into one rank and misses the finer lane ordering.}
\label{fig:qualitative}
\end{figure*}

\paragraph{Results.}
Experiment 2 asks whether training on the attributes is necessary, and
Table~\ref{tab:exp2} answers it: no zero-shot VLM segmenter reproduces the dense
maps. On the CARLA test set, our model outperforms all baselines on both
attributes, with 79.74 / 83.57 drivability/vulnerability mIoU compared with
30.79 / 55.72 for the strongest baseline, LISA.

This gap is not uniform across attributes. All baselines obtain higher mIoU on
vulnerability than on drivability, e.g., 55.72 vs. 30.79 for LISA and 48.23 vs.
3.77 for CorrCLIP. This trend reflects the different reasoning demands of the two
attributes. Vulnerability can often be inferred from local region appearance,
since it mainly asks how severe a collision with that region would be.
Drivability, however, is inherently relational: it depends on lane connectivity,
direction of travel, and the traffic-rule state of the surrounding scene.
Consequently, the baselines fall furthest behind on drivability. The CARLA test column of Figure~\ref{fig:qualitative} makes this visible. 
LISA fragments the drivability map into scattered patches, yet on vulnerability, it still segments the car, the motorcycle, and the buildings. 
OVSeg is worse on both, collapsing the scene into almost a single drivability class and a single vulnerability rank.

The anomaly benchmarks further support this conclusion, although their
drivability metric is less strict. Unlike CARLA, where drivability is evaluated
as a dense scene-level map, SMIYC and StreetHazards evaluate only the labeled
anomaly object and ask whether it is predicted as non-drivable. Our model reliably identifies novel objects as non-drivable,
achieving 98.63\% on SMIYC and 99.11\% on StreetHazards. But most baselines fail this check, with CorrCLIP, F-LMM, and PSALM below 30\% recall on both anomaly sets. Only LISA approaches our performance (69.43 / 91.57). This is a safety-relevant failure because those models often leave anomalous objects drivable.

A second trend appears across model families. The strongest baseline in each
column is a referring segmentation method, whereas the CLIP-based open-vocabulary
methods never achieve the best result. This is consistent with their different
mechanisms: CLIP-based models primarily rely on image-text embedding similarity,
which provides limited support for relational reasoning. Referring methods can
leverage generative VLM reasoning and therefore perform better, but they still
remain below our model.

The question here is not which model segments better in general, but whether
these maps can be produced by prompting alone. They cannot. Prompting existing
VLM segmenters is insufficient to reproduce the dense attribute maps learned by
our model.

\section{Ablation}
\label{sec:ablation}

\subsection{Tapped layer}
\label{ablation:tapped layer}

\begin{figure}[htbp]
  \centering
  \includegraphics[width=0.9\columnwidth]{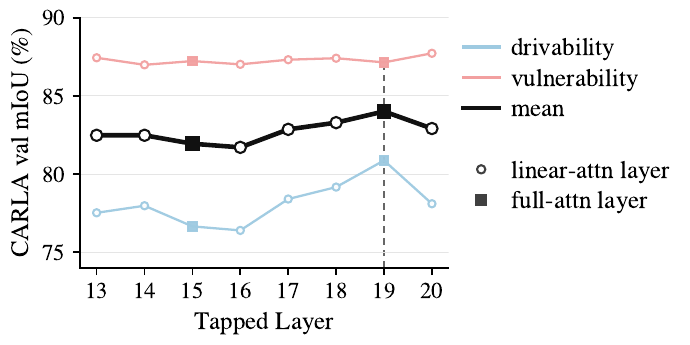}
  \caption{\textbf{Tapped Qwen layer.} CARLA val mIoU when dense features are read from different Qwen layers. Layer~19 gives the best mean mIoU. Vulnerability is stable across layers, while drivability depends more on depth. Marker shape indicates the tapped layer's attention type.}
  \label{fig:tap_layer}
\end{figure}

We read dense features from a single layer of Qwen, which makes the tapped layer a design choice.
Skean \emph{et al.}~\cite{skean2025layer} show that the intermediate layers of a language model encode its most informative and transferable features, whereas the final layers specialize to next-token prediction and transfer less well.
Motivated by this, we sweep the read-out across the middle of Qwen's 32-layer stack, from layer~13 to~20, retraining the model at each depth with all other settings fixed (Fig.~\ref{fig:tap_layer}).
The two attributes depend on depth in different ways.
Vulnerability changes little, whereas drivability is more depth-sensitive.
The mean mIoU peaks at layer~19, which lies at about $0.6$ relative depth, and we use this layer in all experiments.

\subsection{PointRend}
\label{ablation:pointrend}

\begin{figure}[htbp]
  \centering
  \includegraphics[width=0.9\columnwidth]{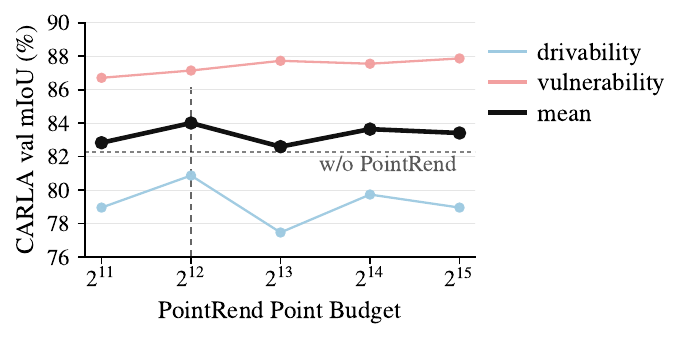}
  \caption{\textbf{PointRend refinement budget.}
  CARLA val mIoU across point budgets. The dashed line is bilinear upsampling without PointRend. The best mean mIoU is obtained with $2^{12}$ points.}
  \label{fig:pointrend}
\end{figure}

The decoder uses PointRend~\cite{kirillov2020pointrend} in the final two upsampling stages.
At each stage, PointRend samples pixels with high prediction uncertainty and re-predicts them at a higher spatial resolution.
Thus, both the use of PointRend and the number of sampled points are design choices.
We first remove PointRend entirely, replacing the refinement stages with bilinear upsampling, and then sweep the point budget from $2^{11}$ to $2^{15}$ (Fig.~\ref{fig:pointrend}).
PointRend improves the mean mIoU at every tested budget, with gains over the bilinear baseline ranging from $0.30$ points at $2^{13}$ to $1.71$ points at $2^{12}$.
For vulnerability, the gain is steady and saturates beyond $2^{13}$, whereas drivability varies more across budgets.
The asymmetric gain is expected because vulnerability is more directly tied to visual
boundary cues in the RGB image, while drivability depends more on contextual and relational evidence. 
We set the PointRend budget to $2^{12}$ points, which gives the highest mean mIoU in this sweep.

\subsection{Loss}
\label{ablation:loss}

Our attribute labels are ordered: drivability ranges from non-drivable regions to the ego lane, and vulnerability ranges from inert background to people. This makes ordinal losses a natural baseline. We compare
CORAL~\cite{cao2020coral} and CORN~\cite{shi2023corn} with the sigmoid focal loss used in our model (Tab.~\ref{tab:loss}). We also report rank mean absolute error (rank-MAE), which is the average absolute difference between the predicted and ground-truth rank indices over all valid pixels.
Although ordinal losses explicitly model the rank structure, they do not improve rank-MAE, and they give lower mIoU on both attributes. We therefore use sigmoid focal loss in all experiments.

\begin{table}[htbp]
  \centering
  \caption{\textbf{Loss ablation.} CARLA val mIoU and rank-MAE under different training losses. Sigmoid focal loss improves mIoU over the ordinal losses without increasing rank-MAE.}
  \label{tab:loss}  
  \resizebox{0.8\columnwidth}{!}{%
  \begin{tabular}{@{}lcccc@{}}
    \toprule
     & \multicolumn{2}{c}{mIoU $\uparrow$} & \multicolumn{2}{c}{rank-MAE $\downarrow$} \\
    \cmidrule(lr){2-3} \cmidrule(lr){4-5}
    Loss & Driv. & Vuln. & Driv. & Vuln. \\
    \midrule
    CORAL & 64.77 & 81.49 & 0.2548 & 0.0324 \\
    CORN  & 78.53 & 83.95 & 0.2251 & 0.0299 \\
    Sigmoid focal (ours) & \textbf{80.87} & \textbf{87.15} & \textbf{0.2250} & \textbf{0.0298} \\
    \bottomrule
  \end{tabular}%
  }
\end{table}

\section{Limitation and Conclusion}
\label{sec:conclusion}
\paragraph{Limitation}

VOLA predicts how each region should be treated, but it does not yet \textit{act} on those predictions. Future work can close this loop by coupling dense attribute prediction with planning, so that predicted region properties directly shape driving
decisions and improve the robustness of open-world driving.

\paragraph{Conclusion}
We recast open-world driving perception as dense attribute prediction.
Instead of naming what is in a scene, we predict how each pixel region should be treated.
We instantiate this idea with two safety-relevant attributes: drivability and vulnerability.
The key idea is to read this knowledge directly from a VLM by tapping a single intermediate layer and use its image tokens as a dense view of the scene, with no SAM-style mask model, no added special tokens, and no autoregressive text generation.
Each image token summarizes a $32\times32$ patch, so the semantics are present but fine spatial detail is lost.
A lightweight decoder restores it from the image and produces sharp full-resolution maps.
Experiments show that VLM image tokens form a strong dense backbone for open-world attribute prediction, while attribute-specific training is needed for producing reliable maps.

{
    \small
    \bibliographystyle{unsrtnat}
    \bibliography{main}
}

\clearpage
\appendix

\section{Implementation Details}
\label{supp:impl}

Table~\ref{tab:impl} shows the configuration of our model and its training.

\begin{table}[htbp]
  \centering
  \small
  \setlength{\tabcolsep}{4pt}
  \renewcommand{\arraystretch}{1.08}
  \caption{\textbf{Configuration} from the run used for the main results.}
  \label{tab:impl}
  \begin{tabular}{@{}>{\raggedright\arraybackslash}p{0.40\columnwidth}>{\raggedright\arraybackslash}p{0.52\columnwidth}@{}}
    \toprule
    \multicolumn{2}{@{}l@{}}{\textit{Backbone and read-out}} \\
    \midrule
    Backbone & Qwen3.5-4B \\
    LM layers & 32 \\
    Tapped layer & 19 $(\mathrm{round}(0.6{\times}32))$ \\
    Patch / merge & $16{\times}16$ / $2{\times}2$ \\
    Token grid & ${\approx}\,1/32$ of input \\
    \addlinespace
    \multicolumn{2}{@{}l@{}}{\textit{Decoder}} \\
    \midrule
    Decoder width & 256 \\
    Dense upsample & 3 stages ($1/32{\to}1/4$) \\
    RGB skip width & 64 (zero-init fusion) \\
    Fine-feature width & 128 \\
    Attribute heads & $1{\times}1$, 7 driv / 5 vuln \\
    PointRend stages & 2 stages ($1/4{\to}1/1$) \\
    Points per stage & $2^{12}=4096$ \\
    Point sampling & oversample 3, importance 0.75 \\
    Decoder dropout & 0.1 \\
    \addlinespace
    \multicolumn{2}{@{}l@{}}{\textit{Optimization}} \\
    \midrule
    Epochs & 12 \\
    Batch size & 4 \\
    LoRA LR & $1{\times}10^{-4}$ \\
    Decoder LR & $5{\times}10^{-4}$ \\
    Schedule & 3\% warmup, cosine, min $0.1{\times}$ \\
    Weight decay & 0.01 \\
    Loss & sigmoid focal ($\gamma{=}2.0$, $\alpha{=}0.25$) \\
    \bottomrule
  \end{tabular}
\end{table}

\newpage
\section{Prompt Fed to the VLM}
\label{supp:prompt}

As described in Sec.~3.2 of the main paper, we place a short text prompt before
the image so that the image-token hidden states become prompt-conditioned
through causal attention. The exact strings are below.

\paragraph{System prompt.}
\begin{quote}\itshape
You are a perception system for autonomous driving. Examine the image carefully
before answering.
\end{quote}

\paragraph{User prompt.}
\begin{quote}\itshape
For every region of this image, think about these two questions:

\noindent 1. Drivability. How safe is it for the ego vehicle to drive across
this region? Consider whether the surface supports motion, whether the path is
obstructed, and whether crossing it would endanger other agents.

\noindent 2. Vulnerability. How harmful would a collision be for whatever
occupies this region? Consider how badly the thing or person there would be
damaged.

\noindent Your reasoning should be about the attributes of each region rather
than the category of what's there. Two regions with similar attributes should
get similar answers.
\end{quote}

\newpage
\section{Cityscapes Label Mapping}
\label{supp:cityscapes}

Cityscapes has no lane-connectivity or traffic-rule annotations, so we cannot
build the full 7-rank drivability map. We therefore collapse drivability into
three groups, non-drivable, off-road, and on-road, and keep vulnerability at its
five ranks. Tables~\ref{tab:city-vuln} and~\ref{tab:city-driv} give the mapping
from Cityscapes label ids to the two axes. For scoring, we collapse our model's
seven drivability ranks to the same three groups: every lane rank (current,
reachable, not-reachable, opposite, and red-light blocked) counts as on-road,
the emergency off-road rank as off-road, and rank~0 as non-drivable. \textit{The
Cityscapes void ids (unlabeled, ego vehicle, rectification border, out of roi,
static, and dynamic) are ignored on both axes.}

\begin{table}[htbp]
  \centering
  \small
  \setlength{\tabcolsep}{4pt}
  \caption{\textbf{Cityscapes vulnerability mapping.} Cityscapes classes grouped
  into our five vulnerability ranks.}
  \label{tab:city-vuln}
  \begin{tabular}{@{}cl>{\raggedright\arraybackslash}p{0.52\columnwidth}@{}}
    \toprule
    Rank & Name & Cityscapes classes \\
    \midrule
    4 & biologicals & person, rider \\
    3 & vehicles & car, truck, bus, caravan, trailer, train, motorcycle, bicycle \\
    2 & walls & building, wall, bridge, tunnel \\
    1 & obstacles & fence, guard rail, pole, pole group, traffic light, traffic sign, vegetation, terrain \\
    0 & non-vuln. & ground, road, sidewalk, parking, rail track, sky \\
    \bottomrule
  \end{tabular}
\end{table}

\begin{table}[htbp]
  \centering
  \small
  \setlength{\tabcolsep}{4pt}
  \caption{\textbf{Cityscapes drivability mapping.} Cityscapes classes grouped
  into the three drivability groups.}
  \label{tab:city-driv}
  \begin{tabular}{@{}l>{\raggedright\arraybackslash}p{0.60\columnwidth}@{}}
    \toprule
    Group & Cityscapes classes \\
    \midrule
    on-road & road \\
    off-road & ground, sidewalk, parking, rail track, terrain \\
    non-drivable & all other labeled classes (building, wall, fence, guard rail, bridge, tunnel, pole, traffic light, traffic sign, vegetation, sky, and all people and vehicles) \\
    \bottomrule
  \end{tabular}
\end{table}

\newpage
\section{Additional Qualitative Results}
\label{supp:qual}

\begin{figure*}[htbp]
\centering
\includegraphics[width=0.75\textwidth]{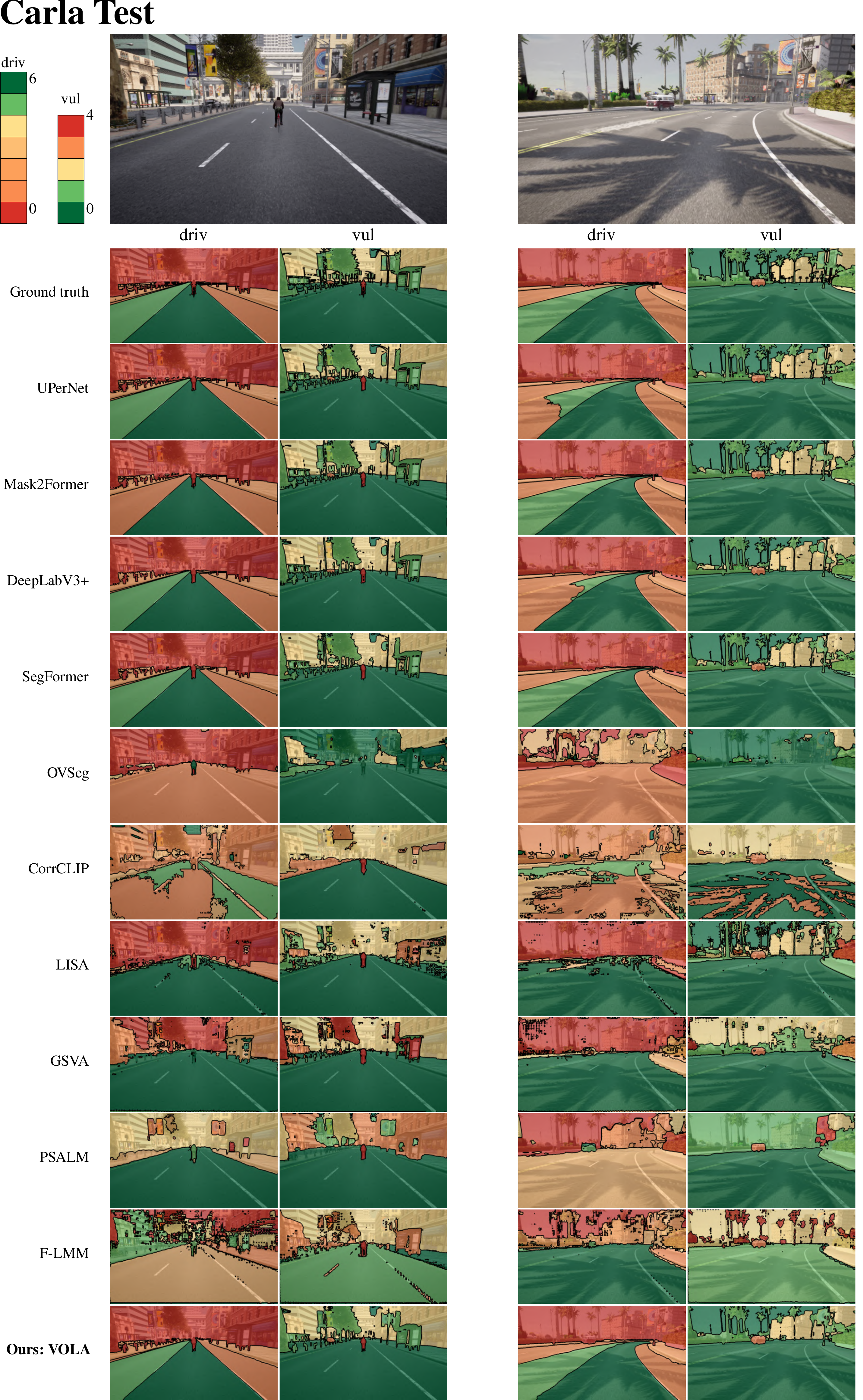}
\caption{Additional qualitative results on CARLA Test.}
\label{fig:supp_carla}
\end{figure*}

\begin{figure*}[htbp]
\centering
\includegraphics[width=0.75\textwidth]{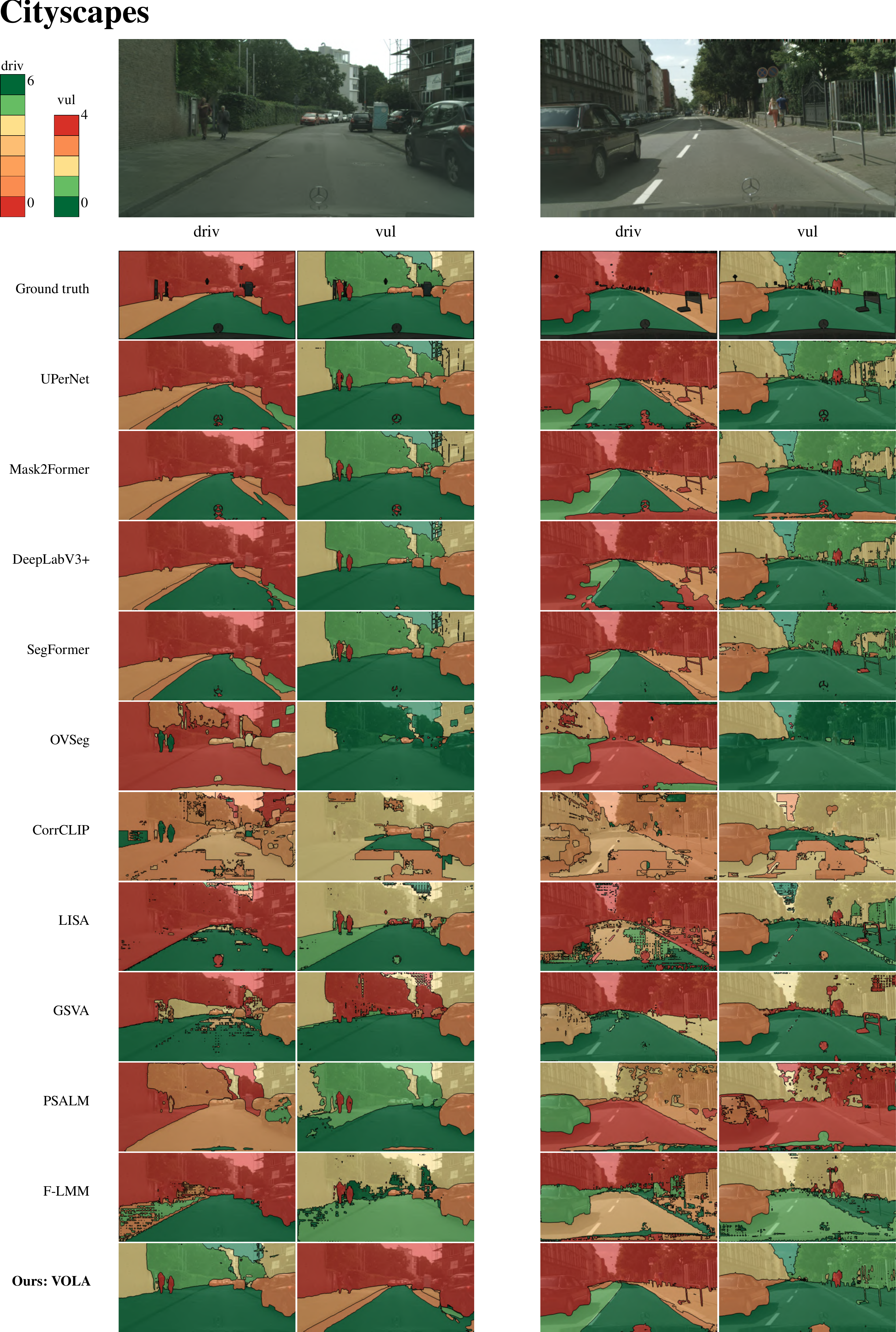}
\caption{Additional qualitative results on Cityscapes.}
\label{fig:supp_cityscapes}
\end{figure*}

\begin{figure*}[htbp]
\centering
\includegraphics[width=0.75\textwidth]{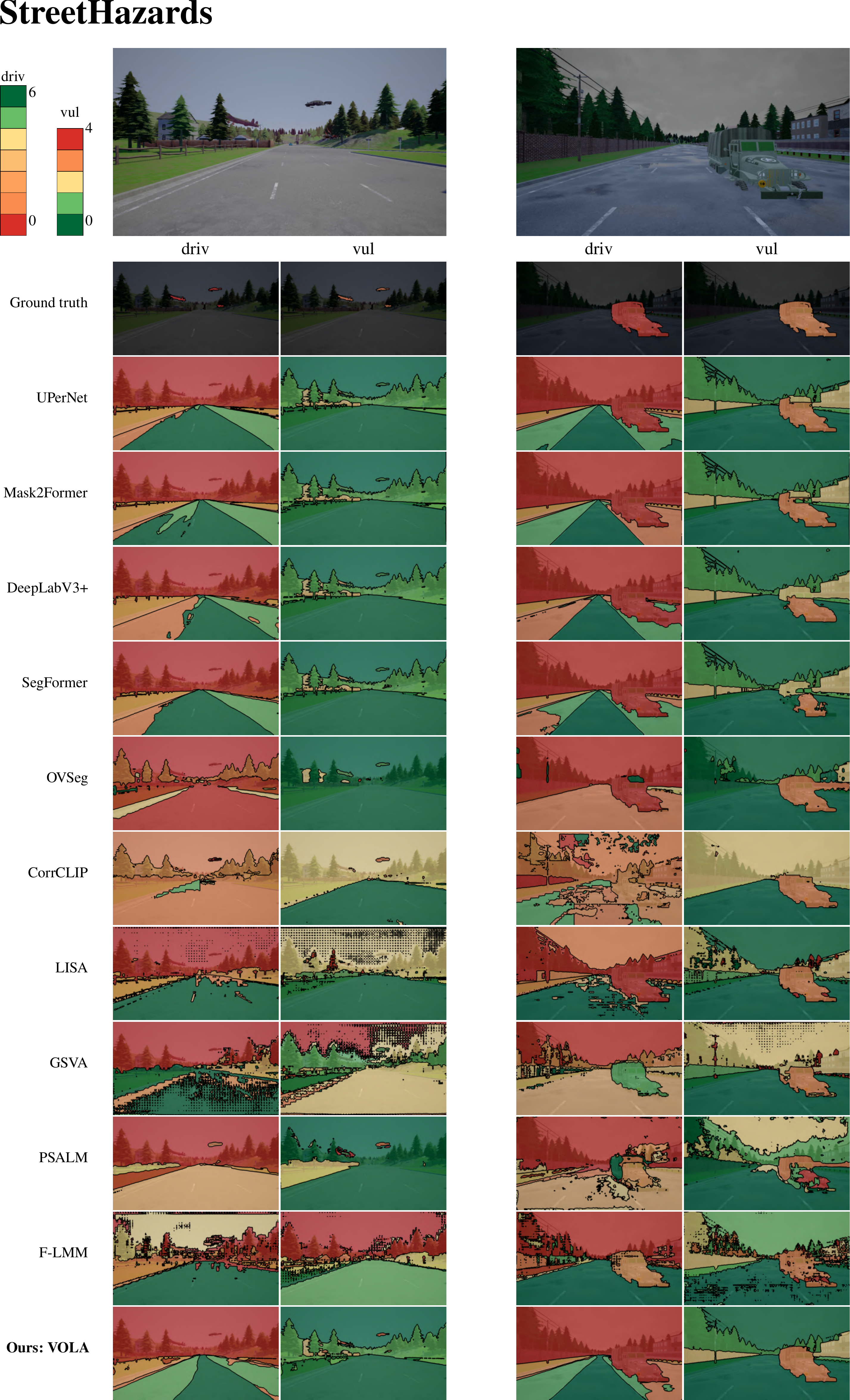}
\caption{Additional qualitative results on StreetHazards.}
\label{fig:supp_sh}
\end{figure*}

\begin{figure*}[htbp]
\centering
\includegraphics[width=0.75\textwidth]{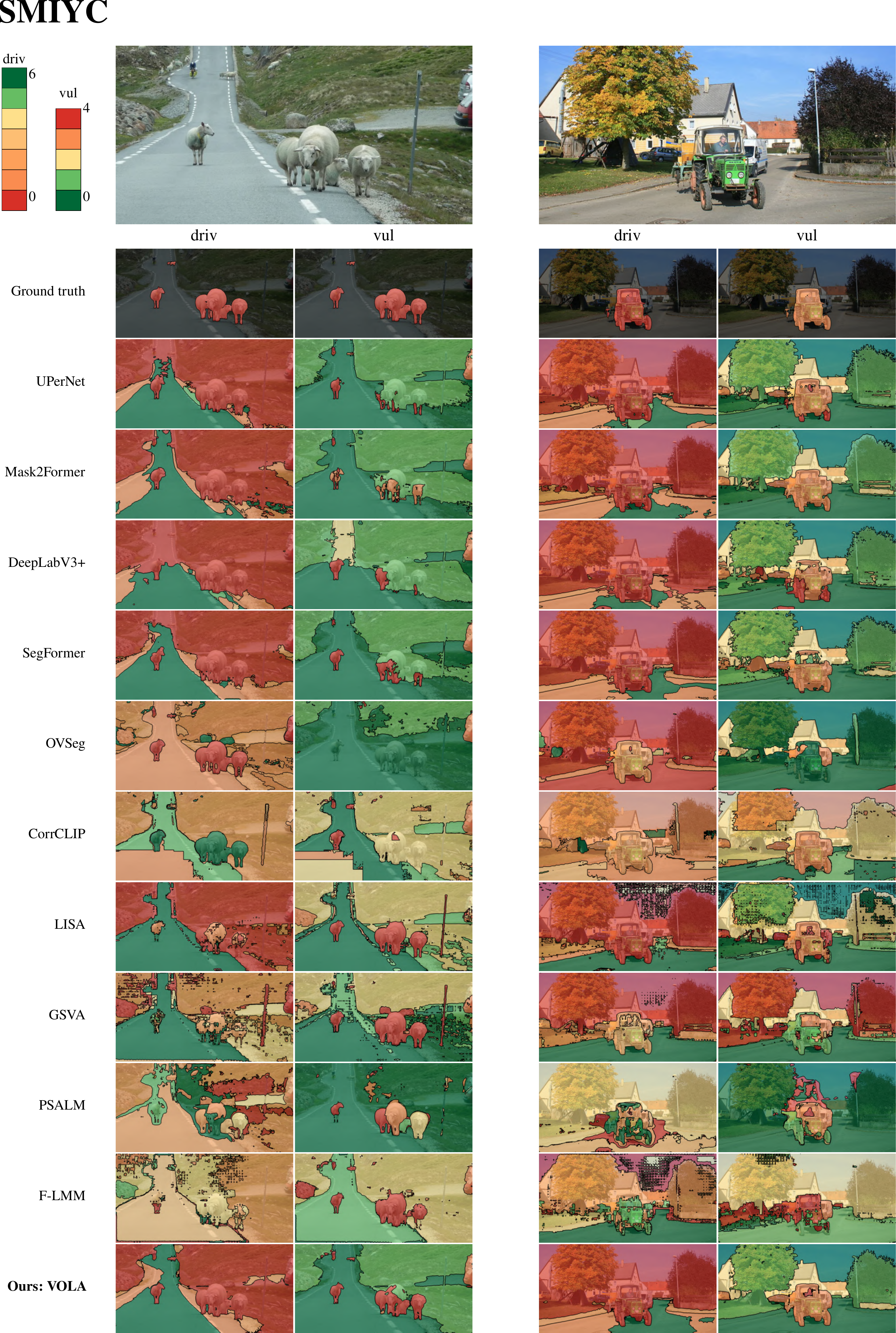}
\caption{Additional qualitative results on SMIYC AnomalyTrack.}
\label{fig:supp_smiyc}
\end{figure*}

The main paper shows one qualitative figure with a subset of strongest methods. Here we
give larger panels with all ten baseline methods so every baseline can be read on the
same input. Figures~\ref{fig:supp_carla}--\ref{fig:supp_smiyc} cover the four
datasets along our distribution-shift gradient: CARLA Test, Cityscapes,
StreetHazards, and SMIYC. In each panel the input image is on top, followed by
one row per method, and for every method the drivability map (driv) and the
vulnerability map (vul) are shown side by side.

\end{document}